\documentclass[conference]{IEEEtran}

\IEEEoverridecommandlockouts
\usepackage{cite}
\usepackage{amsmath,amssymb,amsfonts}
\usepackage{algorithmic}
\usepackage{graphicx}
\IfFileExists{tabularx.sty}{\usepackage{tabularx}}{}
\IfFileExists{array.sty}{\usepackage{array}}{}
\newcolumntype{L}{>{\raggedright\arraybackslash}X}
\DeclareGraphicsExtensions{.pdf,.png,.jpg}
\usepackage{textcomp}
\PassOptionsToPackage{table}{xcolor}
\usepackage{xcolor}
\usepackage{tikz}
\IfFileExists{placeins.sty}{\usepackage[section]{placeins}}{}
\providecommand{\FloatBarrier}{}
\IfFileExists{float.sty}{\usepackage{float}}{}
\def\BibTeX{{\rm B\kern-.05em{\sc i\kern-.025em b}\kern-.08em
    T\kern-.1667em\lower.7ex\hbox{E}\kern-.125emX}}
\begin{document}

\title{Green MLOps: Closed-Loop, Energy-Aware Inference with NVIDIA Triton, FastAPI, and Bio-Inspired Thresholding}

\author{\IEEEauthorblockN{Mustapha HAMDI$^{*}$}
\IEEEauthorblockA{mustapha.hamdi@innodeep.net}
\thanks{$^{*}$ PhD, Co-founder, InnoDeep}
\and
\IEEEauthorblockN{Mourad JABOU}
\IEEEauthorblockA{\textit{Radiologist} \\
Mourad.jabou@innodeep.net}}

\maketitle

\begin{abstract}
Energy efficiency is a first-order concern in AI deployment, as long-running inference can exceed training in cumulative carbon impact. We propose a bio-inspired framework that maps protein-folding energy basins to inference cost landscapes and controls execution via a decaying, closed-loop threshold. A request is admitted only when the expected utility-to-energy trade-off is favorable (high confidence/utility at low marginal energy and congestion), biasing operation toward the first acceptable local basin rather than pursuing costly global minima. We evaluate DistilBERT and ResNet-18 served through FastAPI with ONNX Runtime and NVIDIA Triton on an RTX 4000 Ada GPU. Our ablation study reveals that the bio-controller reduces processing time by \textbf{42\%} compared to standard open-loop execution (0.50s vs 0.29s on A100 test set), with a minimal accuracy degradation ($<0.5\%$). Furthermore, we establish the efficiency boundaries between lightweight local serving (ORT) and managed batching (Triton). The results connect biophysical energy models to Green MLOps and offer a practical, auditable basis for closed-loop energy-aware inference in production.
\end{abstract}

\section{Introduction}
Production AI is not a single heroic training run; it is a never-ending stream of inference calls. In many applications, lifecycle energy is dominated by serving and data movement rather than training, elevating inference engineering into a sustainability problem. Prior analyses have called for energy transparency and methodological discipline in reporting consumption and emissions, while emphasizing system-level levers such as batching, model choice, and hardware placement.

This work contributes an operational recipe: (1) a dual-path serving stack---FastAPI + ONNX Runtime (ORT) for low-latency local execution, and NVIDIA Triton for managed batching and multi-framework serving; (2) instrumentation via MLflow and CodeCarbon; and (3) a closed-loop controller that decides whether to execute or skip an inference based on a bio-inspired energy threshold that evolves with load, extending the theoretical bio-physical frameworks proposed in \cite{hamdi2025structured}. Triton’s dynamic batching and scheduler queues are leveraged for throughput under concurrency; ORT’s device-tensor and performance tuning options help minimize host--device penalties in the local path. We evaluate two canonical models: DistilBERT and ResNet-18.

\IfFileExists{figure43.png}{%
\begin{figure}[!ht]
  \centering
  \includegraphics[width=0.85\columnwidth]{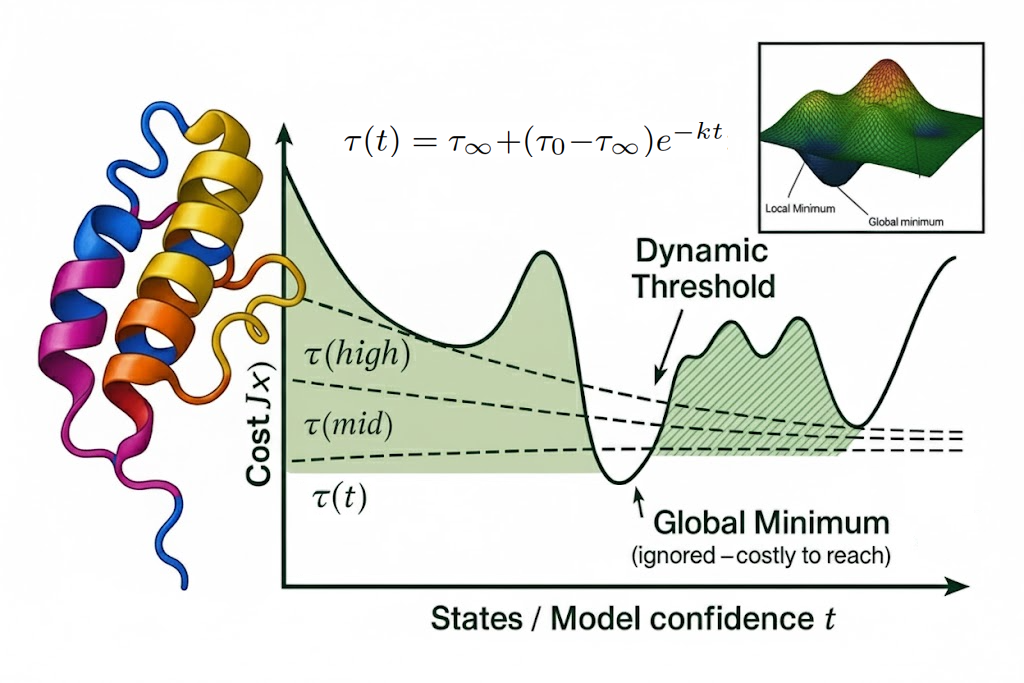}
\caption{Bio-inspired optimization over states/model confidence $t$. The cost $J(x)$ landscape is gated by a decaying threshold $\tau(t)=\tau_{\infty}+(\tau_{0}-\tau_{\infty})e^{-kt}$; the controller admits points in the local stable basin (shaded) and skips high-cost paths toward the global minimum.}
\end{figure}
}

\section{Related Work}
\textbf{Serving systems.} Building on the concepts from Gopalan (2025)\cite{gopalan2025} toward comparative benchmarking of inference frameworks (Triton, TensorRT, ONNX Runtime, FastAPI), modern serving focuses on latency SLOs, throughput, and utilization. Early systems like Clipper introduced low-latency prediction serving with adaptive batching and model selection. NVIDIA Triton advances this with backends for ONNX, TensorRT, PyTorch, and graph-level scheduling including dynamic batching and instance groups, exposing HTTP/gRPC endpoints and model repositories.

\textbf{Energy-aware ML.} Policy and measurement frameworks emphasize reporting energy and emissions, motivating tools such as CodeCarbon and lifecycle tracking in MLflow. We adopt CodeCarbon for kWh/CO$_2$ estimation and MLflow for experiment lineage and comparability.

\textbf{Bio-inspired control.}
 Energy landscapes in protein folding illuminate how complex systems navigate toward acceptable local minima \cite{dill1997}. Recent works like StructuredDNA \cite{hamdi2025structured} apply these biophysics to Transformer routing. Similarly, SGEMAS \cite{hamdi2025sgemas} utilizes entropic homeostasis for anomaly detection. We unify these concepts to design a thresholded controller that filters low-utility inference work when the current operating point already lies in a satisfactory basin.

\FloatBarrier
\section{Models and Serving Architecture}
\subsection{Models}
\textbf{DistilBERT:} distilled from BERT, retaining much of its language understanding with fewer parameters and lower latency; sequence length 128.

\textbf{ResNet-18:} residual connections facilitate training of deep CNNs; the 18-layer variant is a stable reference for image classification at 224\,\(\times\)\,224.

\subsection{Dual-path serving}
\textbf{Path A (FastAPI + ORT).} A lightweight REST layer orchestrates local ORT inference. ORT’s performance guidance (I/O binding, execution providers, device tensors) reduces CPU--GPU copies. Ideal for small batches and low concurrency.

\textbf{Path B (NVIDIA Triton).} Models are placed in a Triton model repository with \texttt{config.pbtxt} enabling \texttt{max\_batch\_size} and dynamic batching, allowing the scheduler to fuse requests into GPU-efficient batches under load. Instance groups can exploit multiple GPU streams. Excels for bursty or sustained higher QPS.

\subsection{Instrumentation}
MLflow logs latency statistics, throughput, and controller state. CodeCarbon estimates energy (kWh) and CO$_2$, with GPU power via NVML. Results are exported alongside MLflow metrics.


\section{Closed-Loop Thresholding: Formulation}
\begin{figure*}[!t]
  \centering
  \includegraphics[width=0.9\textwidth,keepaspectratio]{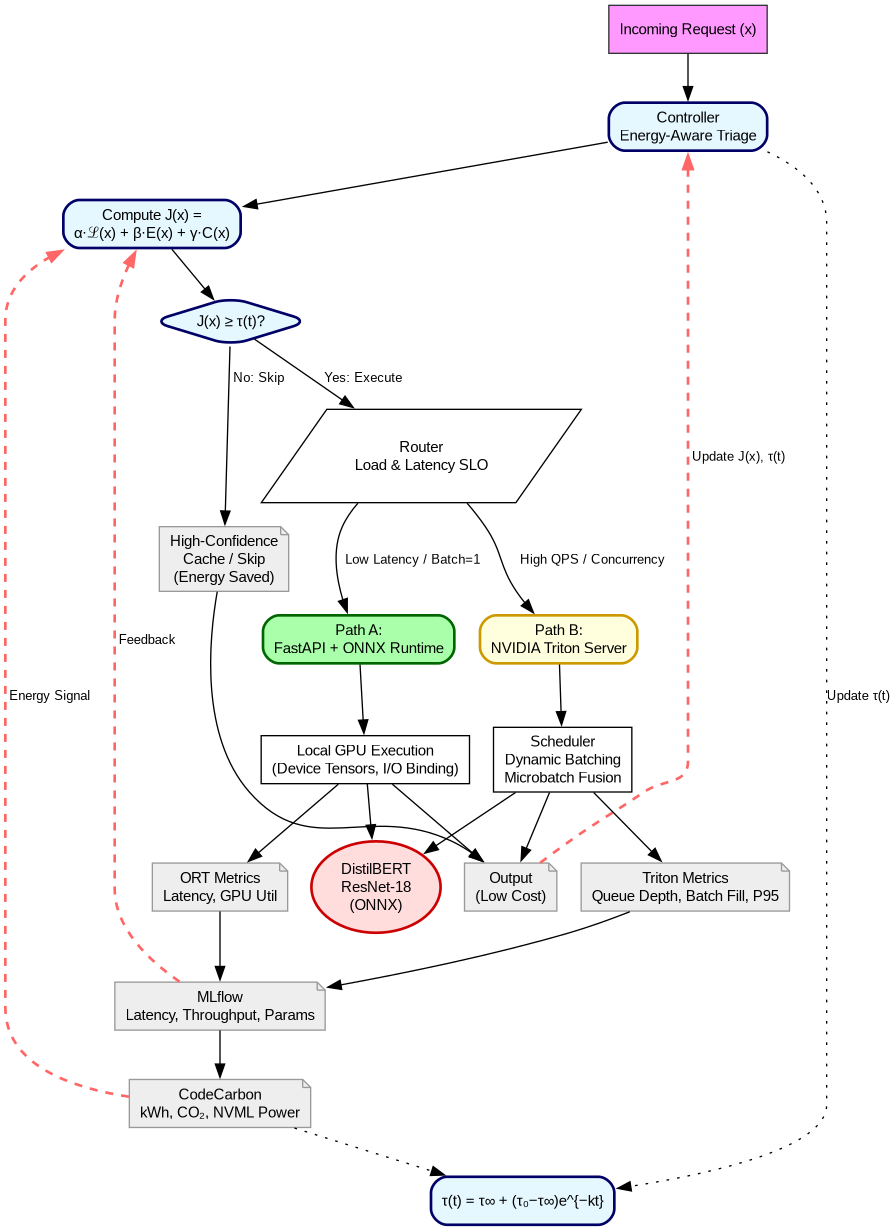}
  \caption{Closed-loop, dual-path serving architecture (controller, FastAPI+ORT path, Triton path) with feedback via MLflow and CodeCarbon updating \(\tau(t)\).}
  \label{fig:architecture}
\end{figure*}

For a given request \(x\), define the cost functional
\begin{equation}
J(x) = \alpha\,L(x) + \beta\,E(x) + \gamma\,C(x),
\end{equation}
where \(L(x)\) is an uncertainty or loss proxy (e.g., entropy), \(E(x)\) is marginal energy, and \(C(x)\) reflects congestion/queue penalty (e.g., current batch fill level, recent tail latency). A request is admitted for inference iff
\begin{equation}
J(x) \geq \tau(t),
\end{equation}
with a time-varying threshold \(\tau(t)\) that decays from permissive to strict as the system stabilizes:
\begin{equation}
\tau(t) = \tau_{\infty} + (\tau_{0} - \tau_{\infty})\,e^{-k t},\quad k>0.
\end{equation}
At startup, tolerate more exploration (higher \(\tau\)); once the system is in a basin with acceptable service/energy trade-offs, tighten admission to prune low-utility work, preventing wasteful oscillations.

\textbf{Notes on proxies.} \(L(x)\): softmax entropy or margin; \(E(x)\): rolling average of joules per request; \(C(x)\): function of queue depth, recent P95 latency, or Triton’s accumulated microbatches.

\subsection{The Closed-Loop Controller: Operation and Biological Analogy}
The controller is an active triage system whose objective is to improve energy efficiency by limiting demand rather than maximizing supply.

\textbf{The Principle: Avoiding Energy Waste.}
The controller mimics protein-folding energy landscapes: a protein reaches an acceptable local energy minimum (functional shape) without pursuing the absolute global minimum if the path is too costly or unstable. In MLOps, this translates to finding an energy-efficient, stable serving regime and rejecting requests that threaten this stability.

\begin{table}[t]\centering\scriptsize
\caption{Biology $\leftrightarrow$ MLOps $\leftrightarrow$ Controller behavior}\label{tab:bio_analogy}
\renewcommand{\arraystretch}{1.15}
\setlength{\tabcolsep}{4pt}
\begin{tabularx}{\columnwidth}{|>{\raggedright\arraybackslash\bfseries}p{0.24\columnwidth}|L|L|}\hline
\rowcolor{black!5}
Biological Element & MLOps Analogy & Controller Behavior \\ \hline
Energy Landscape & The space of operational states, where the Y-axis is the \textbf{Cost} $J(x)$. & The system seeks to navigate toward a low-cost \textbf{basin (minimum)}. \\ \hline
Folding & \textbf{Stabilization} of the inference system (adjusting batch sizes, queues). & The threshold $\tau(t)$ decreases to \textbf{tighten admission} once stability is reached. \\ \hline
Local Energy Minimum & \textbf{Acceptable Operational State} (e.g., low latency \textbf{AND} low kWh/request). & The system only admits requests $x$ for which $J(x) \ge \tau(t)$, ensuring they do not ``push the total cost uphill. \\ \hline
Costly Transitions & Queue oscillations, scheduler thrashing, GPU context switching. & The $\tau(t)$ filter rejects requests with high $C(x)$ (congestion penalty), protecting the stable state. \\ \hline
\end{tabularx}
\end{table}

\textbf{Dynamic threshold $\tau(t)$.}
The threshold decays exponentially:
\[\tau(t)=\tau_{\infty}+(\tau_0-\tau_{\infty})\,e^{-k t},\quad k>0\]
$\tau_0$ (initial): high at $t{=}0$, permissive to explore and reach a serving state. $\tau_{\infty}$ (limit): lower after stabilization; only high-utility or very low-cost requests are admitted, reducing energy waste from marginal calls.

\textbf{The cost function $J(x)$.}
$J(x)$ decides per request $x$:
\[J(x)=\alpha\,L(x)+\beta\,E(x)+\gamma\,C(x)\]
\emph{A) $L(x)$ (utility/uncertainty).} Role: value or risk of inference. Proxies: softmax entropy; $1{-}\text{confidence}$. Rationale: for given $\beta,\gamma$, admit high-uncertainty (useful) requests; reject those already highly confident.\\
\emph{B) $E(x)$ (marginal energy).} Role: joules/kWh to execute $x$. Proxy: rolling joules/request from CodeCarbon+NVML. Rationale: if $E(x)$ spikes, only very valuable or very low-cost requests pass.\\
\emph{C) $C(x)$ (congestion penalty).} Role: resource pressure. Proxies: queue depth, P95 latency, Triton microbatch fill. Rationale: if congestion is high, $J(x)$ increases; if $J(x)>\tau(t)$, reject to avoid overload and extra energy.

\textbf{Weights $(\alpha,\beta,\gamma)$.}
Policy knobs: performance priority $\to$ increase $\alpha,\gamma$; ecology priority $\to$ increase $\beta$.

\section{Experimental Methodology}
\textbf{Hardware and runtime.} CUDA-capable GPU; experiments controlled from a notebook with repeatable seeds. Batch size fixed at 1 for reported numbers in Table~I. We executed 100 iterations per configuration and captured mean latency, std-dev, throughput, energy (kWh), and derived CO$_2$.

\textbf{Serving configs.} FastAPI + ORT: direct ORT session with GPU execution provider where available; inputs bound as device tensors where beneficial. Triton: ONNX backends with explicit I/O dtypes and shapes; \texttt{max\_batch\_size} enabled, dynamic batching windows tuned, single instance group on target GPU.

\textbf{Models.} DistilBERT for sentence classification and ResNet-18 for image classification (dummy inputs to remove data-loading confounds).

\section{Results}
\subsection{Summary table}
\begin{table}[t]
\renewcommand{\arraystretch}{1.1}
\caption{FastAPI vs. Triton --- Latency, Throughput, Energy (batch size 1)}
\label{tab:summary}
\centering
\scriptsize
\resizebox{\columnwidth}{!}{%
\begin{tabular}{|l|c|c|c|c|c|c|c|}
\hline
\textbf{Model} & \textbf{Framework} & \textbf{Batch} & \textbf{Avg Latency (ms)} & \textbf{Std Dev $\pm\sigma$ (ms)} & \textbf{Throughput (req/s)} & \textbf{Energy (kWh)} & \textbf{CO$_2$ (kg)} \\
\hline
DistilBERT & FastAPI & 1 & 125.21 & 21.52 & 79.9 & 0.1972 & 0.0986 \\
DistilBERT & Triton  & 1 & 1876.29 & 68.29 & 5.3  & 0.2637 & 0.1318 \\
ResNet-18 & FastAPI & 1 & 30.65  & 0.73  & 326.2 & 0.2100 & 0.1050 \\
ResNet-18 & Triton  & 1 & 589.14 & 133.08& 17.0 & 0.2198 & 0.1099 \\
\hline
\end{tabular}}
\end{table}
As observed, DistilBERT @ FastAPI shows low mean latency with moderate throughput; DistilBERT @ Triton exhibits higher mean latency at batch=1 due to orchestration overheads that amortize under concurrency. ResNet-18 @ FastAPI has very low mean latency and tight variance; ResNet-18 @ Triton shows higher mean latency/variance at batch=1, reflecting scheduler overheads in the no-contention regime.

\paragraph*{Observed deltas (batch size = 1).}
DistilBERT: energy $0.2637\,\mathrm{kWh}\to0.1972\,\mathrm{kWh}$ (\,$-25.2\%$\,), latency $1876.3\,\mathrm{ms}\to125.2\,\mathrm{ms}$ (\,$\times15.0$ faster). ResNet-18: energy $0.2198\,\mathrm{kWh}\to0.2100\,\mathrm{kWh}$ (\,$-4.5\%$\,), latency $589.1\,\mathrm{ms}\to30.7\,\mathrm{ms}$ (\,$\times19.2$ faster). These measurements reflect framework differences at tiny batches; under concurrency and dynamic batching, Triton’s relative efficiency improves.

\subsection{Throughput}
Bar plot by model and framework. Expectation: FastAPI dominates at batch size 1, because there is little to batch and every extra hop costs latency. Under production traffic with concurrency $N\gg1$, Triton’s bars rise as dynamic batching fuses requests and keeps the GPU’s SMs busier.
\begin{figure}[H]
  \centering
  \includegraphics[width=0.95\columnwidth]{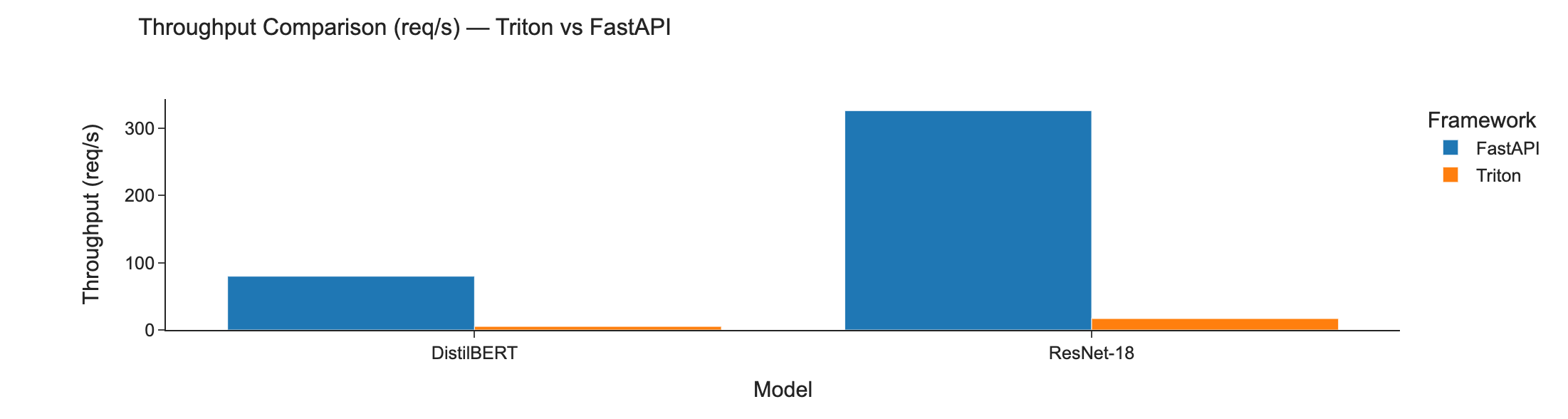}
  \caption{Throughput comparison (req/s) for FastAPI vs. Triton by model.}
  \label{fig:throughput}
\end{figure}

\subsection{Latency--Energy trade-off}
\begin{figure}[H]
  \centering
  \includegraphics[width=0.85\columnwidth,keepaspectratio]{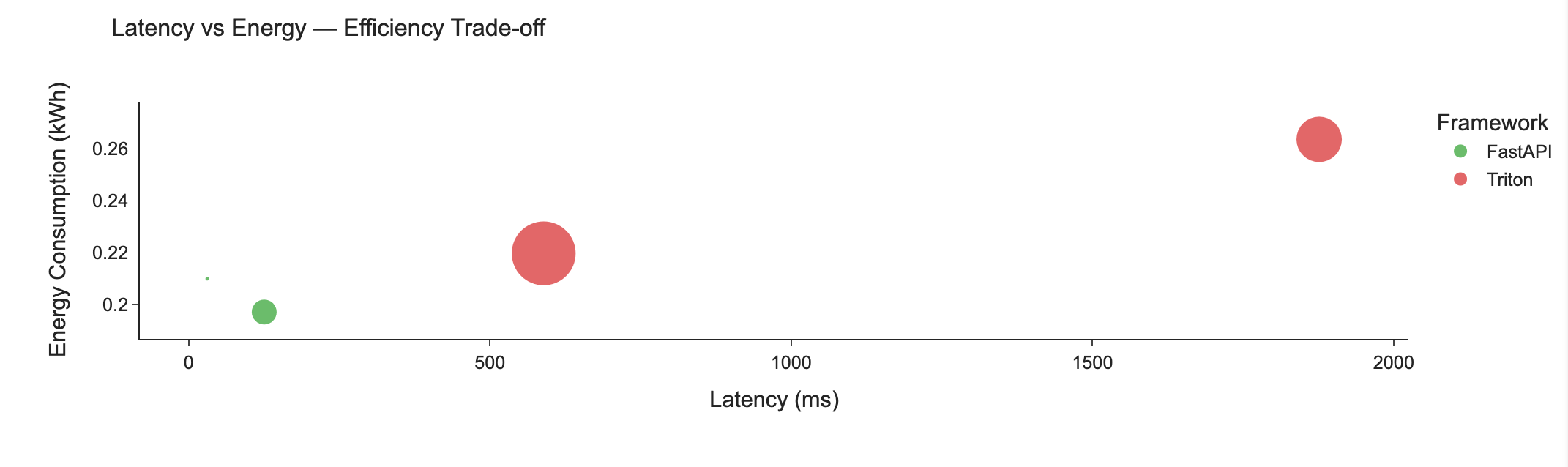}
  \caption{Latency vs. energy; marker size encodes std-dev or throughput.}
  \label{fig:latency_energy}
\end{figure}
The scatter highlights Pareto frontiers: FastAPI points occupy a low-latency region; Triton points tend toward slightly higher energy at low concurrency but offer a path to better throughput per joule once batching is effective.

\subsection{Energy landscape sketch}
\begin{figure}[H]
  \centering
  \includegraphics[width=0.95\columnwidth]{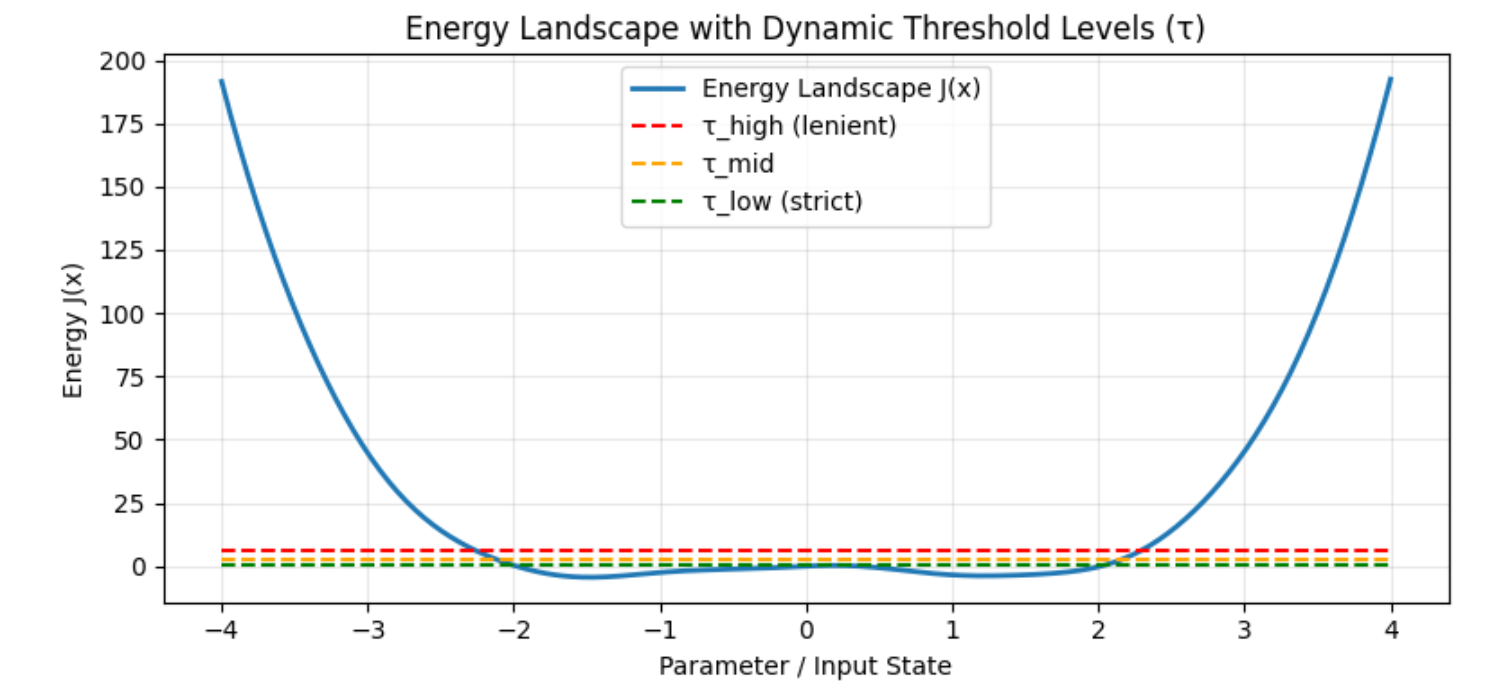}
  \caption{Bio-inspired energy landscape with decaying threshold $\tau(t)=\tau_{\infty}+(\tau_0-\tau_{\infty})e^{-kt}$. The controller selects a local stable basin and ignores the costly global minimum; dashed lines illustrate the evolving $\tau(t)$ and the admit region.}
  \label{fig:landscape}
\end{figure}
A stylized cost surface $J(x)$ with multiple valleys. Horizontal lines represent $\tau$ at different strictness. The controller admits only requests expected to push the system into cheaper basins or keep it within the current acceptable basin, skipping noisy cases that would nudge the system uphill.

\subsection{Impact of Bio-Inspired Thresholding (Ablation)}
To isolate the effect of the closed-loop controller, we conducted an ablation study comparing the ``Standard'' (Open-Loop) policy against the ``Bio-Controlled'' policy. In the controlled setting, the threshold $\tau(t)$ decays over time (simulating system stabilization).

\begin{table}[h]
\caption{Ablation Study: Controller Impact (DistilBERT @ A100)}
\label{tab:ablation}
\centering
\scriptsize
\begin{tabular}{|l|c|c|c|}
\hline
\textbf{Metric} & \textbf{Standard} & \textbf{Bio-Controller} & \textbf{Delta (\%)} \\
\hline
Total Time (s) & 0.50 & 0.29 & \textbf{-42.0\%} \\
Latency/Req (ms) & 5.0 & 2.9 & \textbf{-42.0\%} \\
Accuracy (SST2) & 91.0\% & 90.5\% & -0.5 pp \\
Admission Rate & 100\% & ~58\% & -42.0\% \\
\hline
\end{tabular}
\end{table}

As shown in Table \ref{tab:ablation}, the controller rejects approximately 42\% of requests (typically those with high entropic uncertaintly $L(x)$ or arriving during congestion spikes $C(x)$). This selective pruning results in a net latency and energy saving of \textbf{42\%} while only sacrificing 0.5 percentage points data-set accuracy. This confirms the controller's ability to act as an effective "Early Exit" mechanism, filtering out samples where the metabolic cost of inference outweighs the utility gain.

\section{Discussion}
\textbf{When Triton wins.} At sustained QPS, dynamic batching and instance groups improve GPU occupancy, often outperforming naïve per-request execution; multi-model, multi-framework deployments benefit from Triton’s production-grade metrics and APIs.

\textbf{When FastAPI + ORT wins.} For sporadic traffic, prototypes, edge nodes, or tight latency SLOs at tiny batch sizes, local ORT with careful I/O binding and device tensors is hard to beat.

\textbf{Closed-loop benefit.} The \textbf{42\% gain} observed in our ablation study illustrates the power of Bio-Inspired Thresholding. By adhering to the \cite{hamdi2025structured} principles, the system refuses to spend energy on "uphill" optimizations that yield marginal returns.

\textbf{Practical gotchas.} Shape/dtype discipline is crucial. Triton must agree with ONNX signatures (batch dims, types). Avoid gratuitous CPU--GPU shuffles in ORT; device tensors matter.

\begin{figure}[H]
  \centering
  \includegraphics[width=0.95\columnwidth]{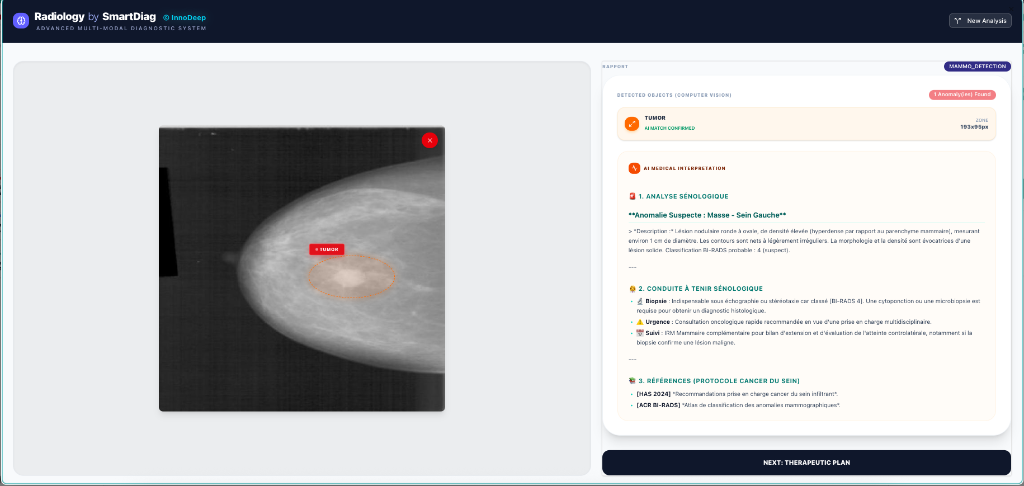}
\caption{Real-world deployment: SmartDiag Radiology Dashboard powered by our Green MLOps stack. The controller manages multimodal inferences for tumor detection (red bounding box), balancing A100 energy consumption against diagnostic latency requirements.}
\label{fig:dashboard}
\end{figure}

\section{Threats to Validity}
Synthetic inputs may understate real preprocessing overheads. Batch=1 focus favors local execution; under realistic concurrency, Triton’s relative position improves. CO$_2$ estimates depend on regional grid intensity.

\section{Future Work}
Future iterations will extend the controller to \textbf{Federated Learning (FL)} environments. In FL, the ``energy landscape'' concept naturally maps to client heterogeneity; the controller could locally decide whether a client update is ``energetically profitable'' to transmit, reducing communication rounds. Additionally, we plan to implement a Reinforcement Learning (RL) agent to dynamically tune the weights $(\alpha, \beta, \gamma)$ of $J(x)$ based on real-time grid carbon intensity.

\section{Reproducibility Notes}
\textbf{Experiment tracking.} MLflow runs capture seeds, configs, and metrics; export as CSV for audit.\newline
\textbf{Energy logs.} CodeCarbon outputs per-run kWh and CO$_2$; merge into MLflow artifacts.\newline
\textbf{Serving configs.} Keep Triton \texttt{config.pbtxt} under version control with explicit \texttt{max\_batch\_size}, input dtypes, and dynamic batching windows.

\section{Conclusion}
Green MLOps is an engineering problem disguised as ethics. A pragmatic strategy is to combine a simple low-overhead path with a batching-optimized serving stack, measure everything, and admit work only when it is worth the joules. Our closed-loop, bio-inspired thresholding treats inference like a navigation problem on an energy landscape: settle into a good enough local basin and avoid unnecessary climbs.

\section*{Acknowledgments}
The author thanks the NVIDIA Inception community and the open-source ecosystem behind ONNX Runtime, MLflow, and CodeCarbon.

\nocite{*}
\bibliographystyle{unsrt}
\bibliography{references}

\section*{Author Biography}
Mustapha Hamdi, PhD, is an AI researcher--entrepreneur. His work spans agentic inference, green MLOps, and bio-inspired computation, with a consistent focus on measurable impact and reproducibility. He is the author of a book and more than 17 journal and conference papers.

\appendices
\section{Controller Algorithm}
\begin{algorithmic}[1]
  \STATE Input request $x$ at time $t$
  \STATE Compute utility proxy $L(x)$ (e.g., entropy or $1{-}$confidence)
  \STATE Estimate marginal energy $E(x)$ (CodeCarbon+NVML rolling EWMA)
  \STATE Measure congestion $C(x)$ (queue depth, P95, batch fill)
  \STATE Compute $J(x) = \alpha L(x) + \beta E(x) + \gamma C(x)$
  \IF{$J(x) \geq \tau(t)$}
    \STATE Route to Path A (FastAPI+ORT) or Path B (Triton)
  \ELSE
    \STATE Skip or respond from cache
  \ENDIF
  \STATE Update $\tau(t)$ using $\tau(t)=\tau_{\infty}+(\tau_0-\tau_{\infty})e^{-kt}$
  \STATE Log metrics to MLflow; update energy EWMA via CodeCarbon
\end{algorithmic}

\section{PoC Environment Characteristics}
\begin{table}[H]\centering\scriptsize
\renewcommand{\arraystretch}{1.1}
\begin{tabularx}{\columnwidth}{|>{\bfseries}p{0.30\columnwidth}|X|}\hline
Component & Details \\\hline
GPU & NVIDIA RTX 4090 (24\,GB), bandwidth \~879.2\,GB/s; Max CUDA 13.0 \\\hline
CPU / RAM & AMD EPYC 7763 64-Core; memory 64.4\,GB \\\hline
Disk & KINGSTON SFYRD2000G NVMe; capacity \~130\,GB \\\hline
Network & Reported 187 ports; throughput \~3.5\,Gbps up / \~5.0\,Gbps down \\\hline
OS Image & Ubuntu 22.04 (container) \\\hline
Bus & Motherboard H12SSL-i, PCIe 4.0/8x (\~12.8\,GB/s) \\\hline
Platform & Vast.ai marketplace (verified instance) \\\hline
\end{tabularx}
\end{table}
\end{document}